# A propagation matting method based on the Local Sampling and KNN Classification with adaptive feature space

Xiao Chen [1, 2)], Fazhi He [1, 2)*]

[1)] (*State Key Laboratory of Software Engineering, Wuhan University, Wuhan, China, 430072*)

[2)] (*School of Computer Science, Wuhan University, Wuhan, china, 430072*)

(fzhe@whu.edu.cn)

Abstract: Closed Form is a propagation based matting algorithm, functioning well on images with good propagation . The deficiency of the Closed Form method is that for complex areas with poor image propagation , such as hole areas or areas of long and narrow structures.    The right results are usually hard to get. On these areas, if certain flags  are provided, it can improve the effects of matting. In this paper, we design a matting algorithm by local sampling and the KNN classifier propagation based matting algorithm. First of all, build the corresponding features space according to the different components of image colors to reduce the influence of overlapping between the foreground and background, and to improve the classification accuracy of KNN classifier. Second, adaptively use local sampling or using local KNN classifier for processing based on the pros and cons of the sample performance of unknown image areas. Finally, based on different treatment methods for the unknown areas, we will use different weight for augmenting constraints to make the treatment more effective. In this paper, by combining qualitative observation and quantitative analysis, we will make evaluation of the experimental results through online standard set of evaluation tests. It shows that on images  with good propagation , this method is as effective as the Closed Form method, while on images in complex regions, it can perform even better than Closed Form.

## 1 Introduction

Image segmentation and matting are hot issues in the field of image processing, both of which are to separate a part of the image from the whole image. The separated parts are called foreground, while the rest are known as the background. The difference is that the result of segmentation is binary; while the result of matting is a continuum of values between 0 and 1. Matting technology is widely used in digital image editing, video effects production, virtual reality and other fields.

Porter and Duffer defined image I as linear combination[1] of the foreground (F) and the background(B) under opacity ($a$).

$$I = \alpha F + (1-\alpha)B \quad (1)$$

In this way, the matting issue can be defined as any point in the image to calculate the values of F and $a$ of that point.

Formula (1) is a problematic equation. As the three parameters on the right side are unknown, therefore if I is a grayscale, then the Formula (1) is an equation with three unknowns, while if I is three-channel color image, then the Formula (1) is an equation with 7 unknowns. Therefore, the matting issue is essentially a problem with no exact solution and has always been a challenging issue in the field of image processing.

In order to solve the matting problem, it's usually need to add some additional constraints manually by the user. Matting with user participation is also known as the interactive matting. There are two kinds of common interaction methods: (1) trimap method: the image is divided into known foreground, background region and unknown area. Trimap is a generic method, but mainly corresponds to the sampling based method.(2) Simple scribe method: use a small number of strokes and some different colors to mark the image pixel as foreground and background respectively, as a basic sample. This method mainly corresponds to the propagation based method.

## 2 Related works

In general, matting algorithm[2] can be divided into three categories: blue screen matting[3], natural image matting and video matting. Blue screen image refers to the images with blue or green background. Natural image refers to images with any background. Mainstream natural image matting can be divided into two categories: the sampling-based method and the Propagation-based method.

The sampling-based method uses the continuity and similarity of the image colors[4-13], assuming that for each of the unknown point, the values of F, B can be estimated by the samples. And then calculate the opacity through the collected samples and the corresponding calculation formula.

Knockout method[4] is an earlier matting algorithm used for local sampling of the boundary areas. The method is simple and fast for calculation, but only effective for smooth image. And it is a non-statistical sampling matting algorithm with no statistical processing for the samples.

Since then, most of the matting algorithm conduct statistical processing on the samples. Ruzon-Tomasi method[5] conducts clustering processing on the samples[14], using Gaussian mixture model to select the foreground and background samples.[6] Hillman method uses principal component analysis to process samples. Bayesian matting[7] assuming that the distribution of the samples is in conformity with the Gaussian distribution, first conducts sample clustering[14], and then uses the maximum posteriori probability to establish the mathematical model for the colors-opacity of the image for the values of the most appropriate foreground and background and the $\alpha$ value of unknown points.

Method based on perceptual color spaces[8] proposed a sample processing method based on color spaces, by transforming the color spaces to reduce the possibility of miscalculation with the same accuracy of matting results and a much faster calculation speed compared with the Bayesian method[7].[5][6][7][8] use the projection method to calculate the $\alpha$, which is also the current common solution with sampling-based method, but at a slower speed. FuzzyMatte[15] calculates the fuzzy degree of connectivity between the unknown and the known points for sampling and calculation. Shared Sampling[9] improves the calculation speed by separate the same background and foreground shared in the images of the region, but requires the image textures to be simple and straightforward.

Local sampling requires that valid samples must exist in the neighboring areas, which can be satisfied in most cases, but there are also a few unknown points with no valid samples existed in the neighboring areas, resulting in unsatisfactory results. Therefore, J. Wang, put forward the Robust Matting[16], to extend the sampling scope from local to global, making matting algorithm no longer dependent on the effectiveness of local samples, thereby largely increasing the stability of the results. Global Matting[10] also uses Global sampling method, and selects the right samples by using random search algorithm, improving the calculation speed. Sample pair method[11] ensures the continuity of the samples by sampling along the gradient direction. GNS method[12] is a global sampling method with no parameters, combining texture information with local texture structure to reduce sample collection errors. SPS Matting[13], built on Global Matting, proposes a new strategy for sample pair selection, using continuous information to improve the effects of the matting. Compared to local sampling, the global sampling largely improves the accuracy of the samples. Shahrian[17] made further improvement on sampling method, determining the scope of the sampling based on the shortest distance between the foreground and background and the unknown points, which no longer confined the sampling range within the edging areas and enriched the sampling scope with more samples collected. The author would conduct clustering processing on the samples before calculations, which will enlarge the sampling pool without lowering the computational speed substantially. Ting method[18] by processing the image depth information as samples, extended of the traditional 3D samples into 4D samples. Method based on fuzzy connection degrees[19] calculated the fuzzy degree of connection to find the strongest paths from the unknown pixels to the foreground and background borders, to collect the adjacent pixels based on the known pixel associated with that path and create a sample set of known and unknown pixels.

Propagation-based matting method using field theory, taking the image as a field[20] with each point only related to points within a certain range, transforms the opacity solving process into the field solving process. The method based on pixel spread can be divided into two categories according to the different generation modes of pixel relations: the method with direct use of pixels and the method using machine learning to get the pixel relations.

BP method[21] using belief propagation , put forward a scribe-based method with a small amount of samples, by solving the relationship between the pixels to get a solution. Easy Matting method[22] deals with the matting problem through iterative energy minimization framework. Spectral matting method[23] calculated the image spectrum to automatically conduct matting processing. Closed Form (CF)[24] using a closed form method introduced the spectrum analysis and matting Laplacian matrix in matting algorithm, assuming that the opacity is approximately linear in the neighborhood to calculate the relations between the pixel and other pixels, using the relations among various points for matting Laplace matrix elements. This method was a breakthrough in the field of matting, after which a variety of matting algorithm based on propagation are focused on the generation of Laplacian matrix elements. Large kernel-based method[25][24] expanded the scope of windows to make the pixels relations more accurate, and adopted block processing to improve the computing speed, while L1 Matting[26]used norms for calculation. The sketch sample processing required that a small amount of samples drew by the user shall represent all the color distribution features, otherwise the final effect of matting will be affected. In this way, Nonlocal[27] (NL) method introduced a new algorithm for reducing the user's input with only a few user inputs required while other works are done by the algorithm, using non-neighborhood principle to calculate the relations between the pixels and dividing samples by clustering algorithm to reduce the user operation while ensuring the effectiveness of the samples and also expanding the relations between the pixels in the neighborhood used by CF to non-neighborhood relations, no longer dependent on the local linear relationship with only disadvantage to be the huge computational overhead. KNN matting used KNN principle instead of the NL principle, to make the samples selected more accurate. CCM method[28] also using the NL principle, put forward a ball model based on color clustering. Simple strokes method[29]specified foreground and background by a few strokes and rectangle dragging. Singaraju method proposed a layer opacity methodology[30]. Fast Closed Form method[31] accelerated CF through hierarchical data structure to strike a balance between the quality and speed of the matting. Scalable Matting method[32], based on such algorithms put forward the method of linear space, improving the processing capacity of large images. Xie method[33] using the active online curve contour model to detect color distribution and then using the level set for partition, is an automatic matting method.

In recent years, Propagation-based method has made new progress, improving the direct computing model used by CF for pixels relations in the neighborhood to using machine learning to calculate relationship between pixels.

Learning Matting (LM) method[34], taking the CF local linear model used to calculate the correlation of pixels as a process of machine Learning, put forward the linear combination expanded as kernel function to reduce dependence on neighborhood data. Peng[35] method built CF Laplace matrix elements by multiple calculation of similarity and multiple iterations, with more accurate description of the relations between unknown points and neighborhood. Transductive approach[36] defined the generation of opacity map as a semi-supervised learning method, followed by methods[37, 38] using support vector regression as a way of classified learning. LNSP method[39] is a learning algorithm combining both neighborhood and non-neighborhood methods, using Robust sampling method as an auxiliary treatment[16]. ITM method[40] used depth map of image for image segmentation and matting processing, and calculated the matting matrix by measuring the similarity. Yang method[41], based on the matting matrix, the neighborhood definition and features space for weighing, put forward a unified framework for digital image matting. Tseng method[42] using matting algorithm for learning based on the multi-level figures, is an unsupervised matting method. Gong[43] using support vector machine (SVM) for foreground segmentation and edge matting, using GPU for algorithm acceleration, greatly improved the processing speed, and applied it to the video processing.[44] Yang method combining face recognition, stroke outline and CF to extract image hair, with high precision of processing results for fine structures such as hair areas.

The sampling-based method can achieve high accuracy of matting with effective samples collected, but the connectivity of the images is relatively poor, while the method based on propagation can achieve results with great connectivity, but not effective in regions where images are difficult to spread. Improving Color Modeling method[45] combines both sampling-based method and Propagation-based method to get a better effect.

## 3 Closed Form matting

Closed Form (CF)[24] matting algorithm is a propagation-based algorithm, taking image as a field with each point only related to the neighboring pixels, first using the similarity among pixels within a certain range to build matting Laplace matrix and then using the existed foreground and background information to build constraint condition for a solution.

$$J(\alpha) = \alpha^T L \alpha + \lambda(\alpha - \beta)^T D(\alpha - \beta) \quad (2)$$

Among which $\alpha$ is the opacity column vector to be solved. L is matting Laplacian matrix. $\lambda$ is the constraint parameter, a larger value, while $\beta$ is the known opacity column vector, among which the foreground is 1, the background and unknown point are zero. D is a diagonal matrix and the diagonal elements are 0 and 1 respectively with the known point corresponding to the value of 1, the unknown point corresponding to 0.

CF matting method is fast with great smoothness of results. However, its disadvantage is that correct results are hard to get for complex areas, such as areas surrounded by foreground or background areas with no background or foreground available samples in the neighborhood, such as hole areas or long and narrow areas. The author puts forward a solution of manually adding some extra constraint information to improve the effects of matting.

In this paper, built on the[24] method, some improvements have been made to improve the treatment in the complicated areas by processing unknown areas by local sampling and KNN classifier with automatically generated additional constraint information.

## 4 Our method

### 4. 1 Pre-processing

Trimap has a limited precision, inevitably leading to some foreground and background points being left out. Therefore, preprocessing can be made on the images and the known areas can be further expanded to reduce the area of unknown regions, so as to reduce the amount of calculating works. Assuming $q \in F$ that the unknown point p satisfies the following conditions, p is deemed as one of the foreground points.

$$(D(p,q) < E_\theta) \wedge (\|I_p - I_q\| \leq (C_\theta - D(p,q))) \quad (3)$$

Among which $D(p,q)$ refers to the space distance between points p and q. $E_\theta$ and $C_\theta$ are space distance threshold and color distance threshold which are empirically set as 9 in our experiments. The processing method for unknown points in the background is similar.

## 4. 2 Additional constraints

The right items of Formula (2) can be divided into two parts, among which $\lambda(\alpha-\beta)^T D(\alpha-\beta)$ is a constraint for $\alpha^T L\alpha$ using the known foreground and background information. For images of simple structure and good propagation, the solution of the Formula (2) can get a good result. But in areas with poor propagation, such as hole areas or long and narrow areas, ideal results are impossible to get, so it is necessary to add more additional constraint information for the unknown regions.

Referencing the second item in the Formula (2), we add the item as follows:

$$\gamma(\alpha - \tilde{\alpha})^T C(\alpha - \tilde{\alpha}) \quad (4)$$

Among which $\gamma$ is the weight parameter column vector, and $\tilde{\alpha}$ is the column vector of the initial value of the unknown area. C is the diagonal matrix, referred to as trust matrix herein, the element corresponds to the foreground and background on the main diagonal is 0, the element corresponds to the unknown point is a reliable value. In this way, Formula (2) is expanded as the following formula (5)

$$J(\alpha) = \alpha^T L\alpha + \lambda(\alpha-\beta)^T D(\alpha-\beta) + \gamma(\alpha - \tilde{\alpha})^T C(\alpha - \tilde{\alpha}) \quad (5)$$

## 4. 3 Initial $\tilde{\alpha}$ and the confidence value

This article uses local sampling and KNN classifier to calculate the additional constraints parameters. K-nearest neighbor method (KNN) is a basic method of classification and regression, among which the input is the feature vectors of samples and the output is samples categories. For new samples, estimation will be made based on its K types of the nearest neighboring training instances under the classification decision-making rules. KNN method, with no learning process, actually uses the training data set for feature vector spaces division. The selection of K, the distance measurement and classification decision-making rules are the three basic elements of KNN.

### 4. 3. 1 Building feature space

In order to improve the accuracy of classification, it is necessary to a feature space of good performance used for classification. Data in CIELAB color space including both the RGB and CMYK color spaces, can preserve the colors in the most effective way. Therefore, this article will establish features in CIELAB color space so as to build a 9D feature space for the input image both color-wise and texture-wise. Among $x_p = \{l_p, a_p, b_p, gx_p^l, gx_p^a, gx_p^b, gy_p^l, gy_p^a, gy_p^b\}$, l, a, b refer to the components in CIELAB color space, the color value range is 0-255, gx and gy are the vertical gradient and horizontal gradient of the image, using gradient value as the image's texture information.

In most cases, the classification by KNN classifier through $x_p$ training will enjoy a high classification accuracy. However, if the image's foreground and background colors are very close, then the classification accuracy by the classifier will be significantly reduced. Therefore, if during the test the accuracy of the classifier is not high, then additional space coordinates will be added as new features herein. Set $x_p = \{l_p, a_p, b_p, gx_p^l, gx_p^a, gx_p^b, gy_p^l, gy_p^a, gy_p^b, coorX_p, coorY_p\}$, in which, $coorX$ and $coorY$ are the space coordinates of P. In case that the image's foreground and background colors are very close, increase space coordinates as parameters can effectively improve the accuracy of classification.

A large number of samples can improve the accuracy of classification, but also can increase the amount of calculation works. This paper uses points on the borders of foreground and background as sample points to reduce the amount of calculation. However, in most cases, sampling of the boundary area is enough to meet the requirements.

## 4. 3. 2 KNN classifier training

Enter the data set: $T = \{(x_1, y_1), (x_2, y_2), ..., (x_n, y_n)\}$, among which $x_i \in \chi \subseteq R^n$ is the feature vector of the example, and $y_i \in Y = \{c_1, c_2, ...c_n\}, i = 1, 2, ..., N$ is the class of the data. Input: the feature vector of the sample $x$. Output: the class of the example $y$.

(1) Determination of k value: Based on the given distance measurements, find K points in the training set T most close to x, taking the neighborhood of x covering the k points as $N_k(x)$.

The selection of K value will substantially affect the classification accuracy of KNN classifier: smaller K value will reduce the approximation error, but very sensitive to the neighbor points, leading to estimation errors if the adjacent points have noise points; whereas larger k value may reduce the estimation errors of learning, yet increase the approximation errors thereof, and may involve those training instances with insufficient similarities into the prediction process, eventually leading to prediction errors.

This article uses the cross-validation method to select the optimal value of k, which is a value with the highest classification accuracy.

(2) Distance metric: The distance between the two examples in the feature space is a reflection of the similarity degree. In this paper, the feature space used is a 9D or 11D vector space $R^n$, n = 9 or n = 11. The similarity will be calculated using Euclidean distance.

$$L_2(x_i, x_j) = \left( \sum_{l=1}^{n} \left\| x_i^{(l)} - x_j^{(l)} \right\|^2 \right)^{\frac{1}{2}} \quad (6)$$

(3) Classification decision-making rules: This article will use the minimum average distance principle to determine x and the category y. The nearest foreground and the background samples will be selected from the sample space to calculate the distance vectors of distF and distB of the unknown point and these points, then calculate each average value respectively with the unknown point classified by the smaller portions of the average.

$$Flag = \begin{cases} 1, & mean(distF) < mean(distB) \\ -1, & otherwise \end{cases} \quad (7)$$

## 4. 3. 3 Calculation of initial value $\alpha$ and the confidence matrix

The initial value $\alpha$ and the trust matrix will be calculated as follows:
F, B, U represent a set of foreground samples, background samples and the unknown area and the matrix of M1, M2 with initial value of 0 respectively.
1) Extract feature space of the sample points, select the appropriate k value and get a KNN classifier with the highest classification accuracy by training.
2) Forecast elements in U with Flag as the prediction result.
3) For $P \in U$, find out the sample points of Fp and Bp nearest to P.
4) Calculate the opacity of the current point:

$$\alpha = \frac{(P - B_p)(F_p - B_p)}{\|F_p - B_p\|^2} \quad (8)$$

5) Calculate the similarity between the current results and the unknown point

$$similarity = \|P - \alpha F - (1-\alpha)B\| \quad (9)$$

6) **if**

$$\begin{aligned} M1(P) &= \exp(-similarity) \\ M2(P) &= \alpha \end{aligned} \quad (10)$$

else
Trust matrix is used to measure the degree of similarity between the unknown point and sample point, and the distance of the feature space will be used herein for the calculation of similarity (10). The distance between the unknown point and sample point gets closer in feature space, the higher the credibility of that point, or otherwise, the lower credibility (11).

$$dis(P) = \begin{cases} \|X_F - X_P\| & Flag(P) = 1 \\ \|X_B - X_P\| & Flag(P) = -1 \end{cases} \quad (11)$$

$$M1(P) = \exp(-\frac{dis(P)}{\sigma^2}) \quad (12)$$

Then this paper uses distance as the initial for calculation. At this time, the dis (P) is a large range of values, and this article uses the Sigmoid function to map it between 0 and 1.

$$M2(P) = \frac{1}{1 + \exp(\rho \frac{-Flag(P)}{dis(P)})} \quad (13)$$

**end**

Flag is the result of classification, among which the foreground is 1, the background is -1, $\varepsilon$ is the judging threshold. $X_F$ and $X_B$ are foreground and background feature samples respectively. This article sets $\sigma^2 = 2$ and $\rho$ as an enlargement coefficient, used to strengthen the effects of classification results and the distance, and this paper sets $\rho = 15$.

C in the M1 corresponding formula (3) and $\alpha$ in the M2 corresponding formula (3), this article sets $\gamma$ as a column vector, foreground and background corresponding element value as 0, the corresponding element values of the unknown point value as 1 or 0. 1. If the corresponding unknown point is calculated by neighboring pixels, the element value is 1, if by the classifier, the element value is 0. 1.

## 5 Experiment and results comparison

Image composition is complicated, therefore will subject to subjectivity only by visual judgment. Rhemann et al in their 2009 ICCV put forward an evaluation method[46] of quantitative analysis, and provided a standard set of test sets through the website www. alphamatting. com. Since then, this method has been widely used in the assessment of matting efficiency[2, 9-13, 17, 25, 27, 28, 31 to 37, 39, 43, 47-50]. Therefore, using this test index and test sets are the most persuasive for the time being.

In this paper, we use all images in the test sets to test the algorithm, and for the evaluation method, the sum of absolute differences (SAD) and mean square error (MSE)are used as evaluation index to evaluate the overall error of matting results. The calculation formula is as follows:

$$SAD = \sum_{i=1}^{N} \left\| \alpha_i - \alpha_i^{true} \right\|_1$$
$$MSE = \frac{\sum_{i=1}^{N} \left\| \alpha_i - \alpha_i^{true} \right\|_2^2}{N} \quad (15)$$

Where $\alpha_i$ is the solving result by the matting algorithm, $\alpha_i^{true}$ is the real results provided by the test sets and N is the total number of pixels.

### 5. 1 Quantitative analysis

In this paper, all the test images are compared with that of CF method.

First, we compare the 07 set in the test sets, which shows among all eight images, except of the 5th image with a worse SAD evaluation result than the CF method, other images all get the best evaluation results. And for the MSE evaluation, the method used herein gets the best performance evaluation.

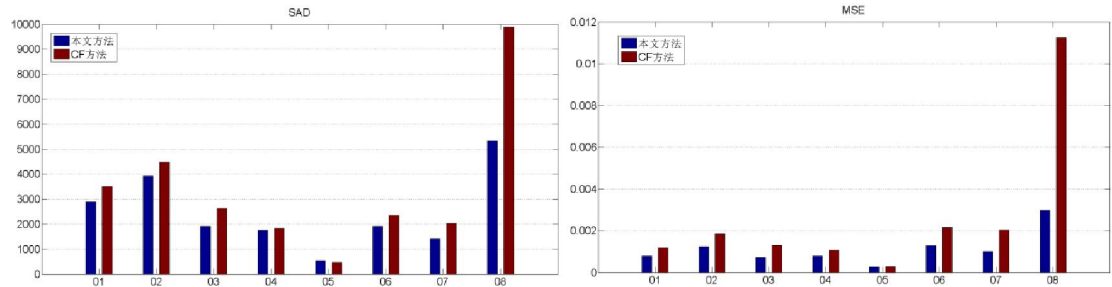

Figure (1) Comparison of test set 07

In the comparison of test set 09, for the evaluation of SAD on the image 8 and 17, CF method has an advantage, while for the rest 25 images, the method herein gets the optimal evaluation results.

For the MSE evaluation on the 17th image, CF method has an advantage, while for the rest 26 images, the method herein gets the optimal evaluation results.

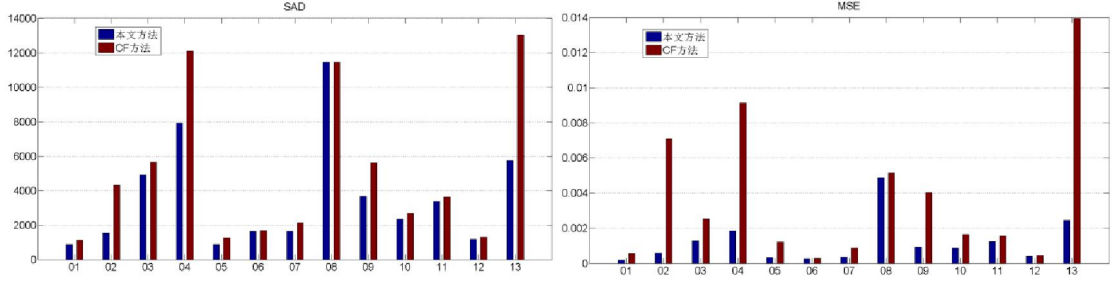

Figure (2) Comparison of images 1-13 in the test set 09

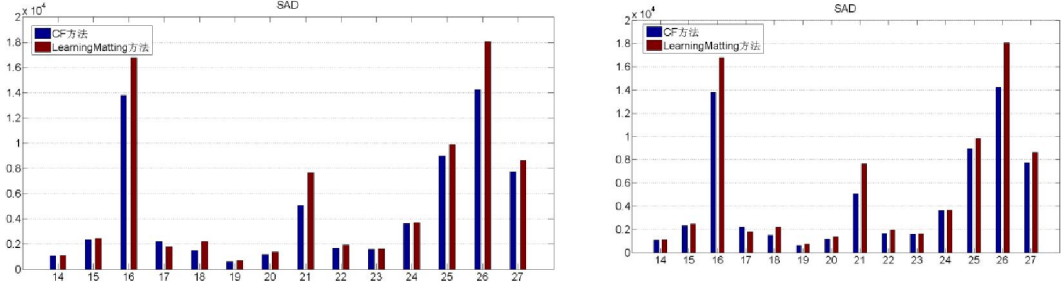

Figure (2) Comparison of images 14-27 in the test set 09

## 5. 2 Qualitative observation

### 5. 2. 1 Comparison of local details

Method used in this paper makes improvements on the limitations of the CF method in complex areas. Figure (5) shows a comparison of local details in two typical images in the 09 test set. Image 4 features a huge unknown area, and the red box highlights a large, long and narrow area with only foreground samples and no background samples nearby, leading to bad image propagation in the region. As all information received by propagation are foreground information, therefore the CF method technically can't get good results in the area with almost all results achieved by CF method therein as the calculation for the foreground, while the results obtained by our method in the area are excellent.

Image 5 features a hole area highlighted by the red box. In the above annular section is an area with poor propagation , but as both neighboring foreground and background samples can be collected, so CF method can be somewhat effective with certain results in this area. However, the following hole area is completely surrounded by the foreground, which can not be processed by the CF method, whereas for our method, this is not a problem and we achieved great results in this area by the combination of both neighboring and global information and by the classifier's processing of the unknown area.

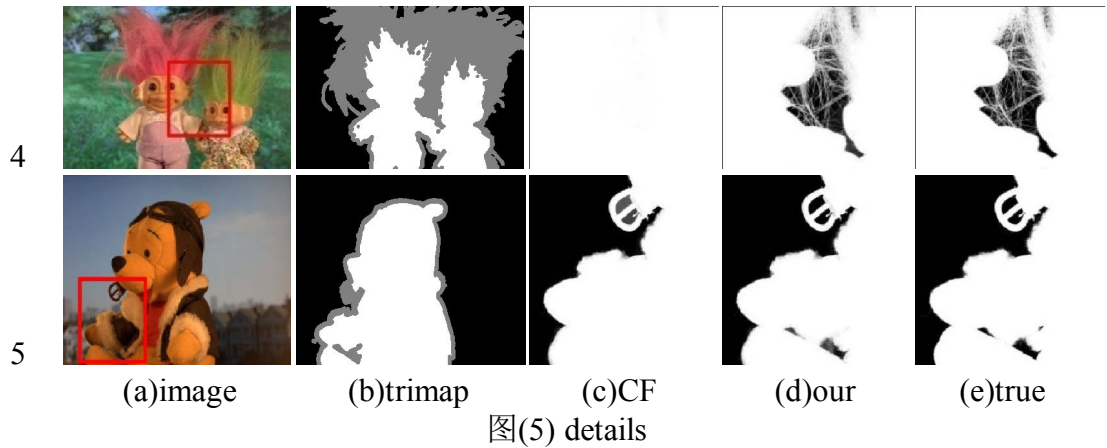

|   | (a)image | (b)trimap | (c)CF | (d)our | (e)true |

图(5) details

## 5. 2. 2 Image comparison

From the image comparison, we can see that if the image composition is simple without too many holes or long and narrow structures, or if the unknown area, though complex, has abundant available neighboring foreground and background samples, then the algorithm used herein has no prominent advantages- see image 17. Otherwise, if the image composition is more complex with holes and lack of enough samples -see images 2, 3, 5, 13, 21, 25, or there are areas of long and narrow structure and lack of abundant effective samples around - see images 4, 9, 18, then the algorithm used herein has obvious advantages in terms of matting results, especially effective for areas difficult to process by the CF method with the details of the complex area preserved well.

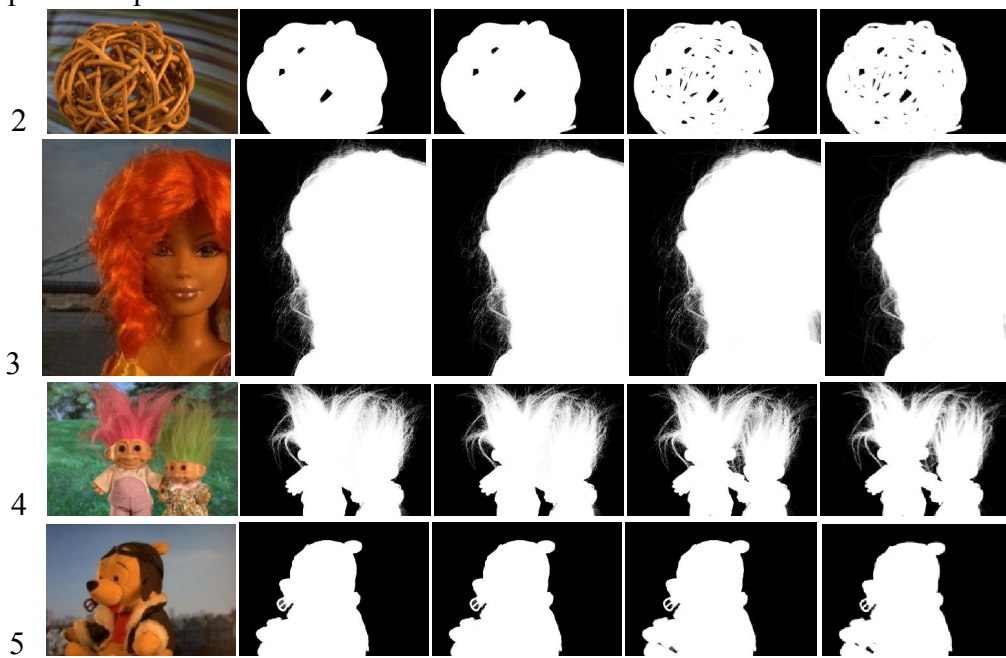

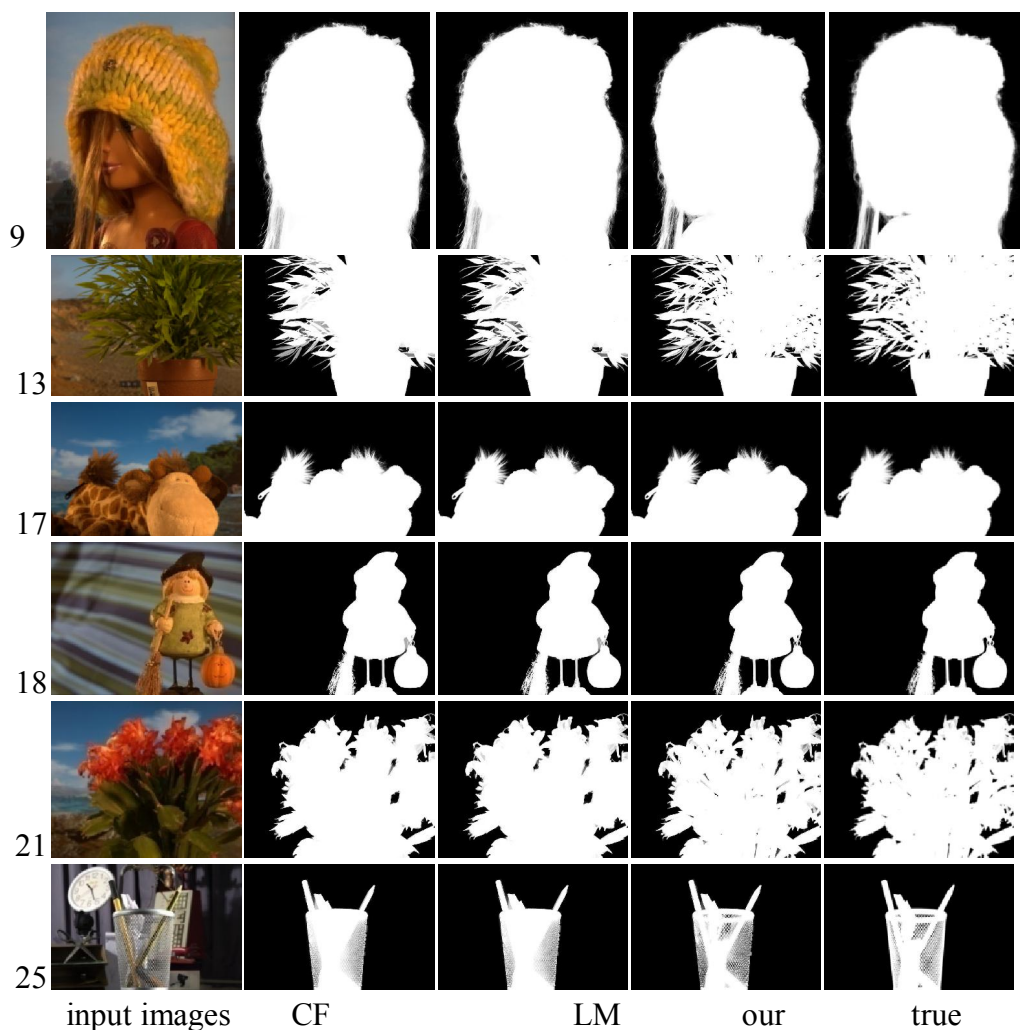

| | input images | CF | LM | our | true |

Figure (6) Comparison of several images

### 5. 3 Matting effects of other images

On www. alphamatting. com, there are a set of images with no accurate results in addition to the standard 07, 09 test atlas. Any matting method's results can be uploaded to the site for evaluation, and the results will be compared with other algorithm of matting. Review will be carried out in three categories: small, large and user, representing the matting results of trimap from accurate to rough in terms of accuracy. In figure (7), the overall shows the overall ranking, while avg. Small/large/user represent three different trimap. Several other algorithms in comparison are KNN matting, learning matting, closed form matting and CF method improvement large kernel matting.

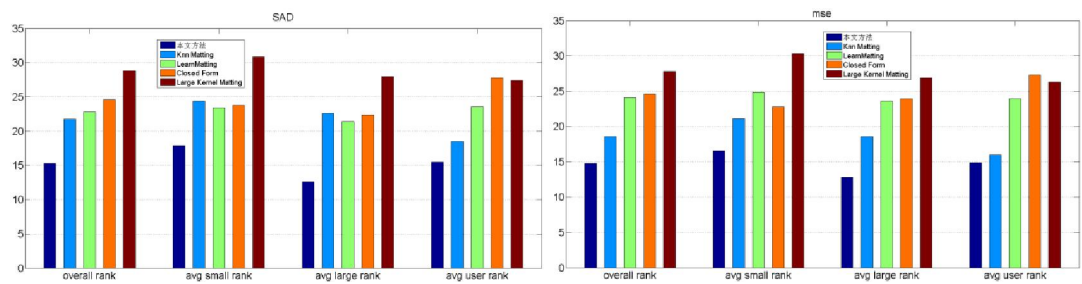

Figure (7) The online evaluation results of our method by www. alphamatting. com

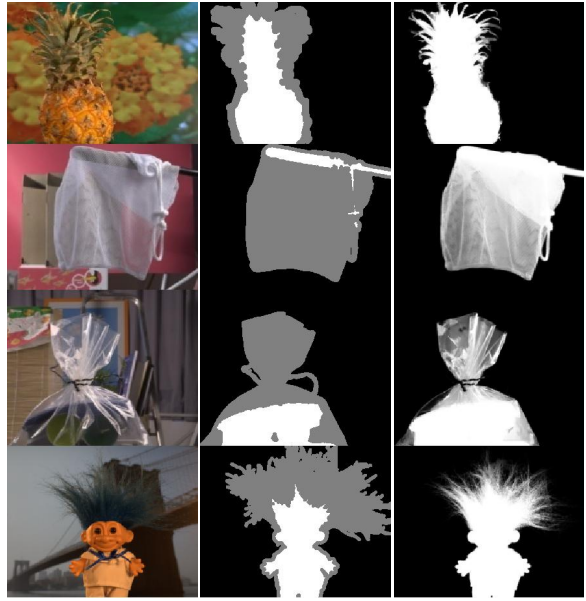

Figure(8) Results of several images evaluated

| SVR Matting | 13.1 | 15.9 | 12.6 | 10.9 |
|---|---|---|---|---|
| Comprehensive Weighted Color and Texture | 13.1 | 13.6 | 13.3 | 12.5 |
| Sparse coded matting | 13.6 | 16.9 | 13.9 | 10.1 |
| **LocalSamplingAndKnnClassification** | **15.3** | **17.9** | **12.6** | **15.5** |
| CCM | 15.4 | 18.3 | 15 | 12.9 |
| Weighted Color and Texture Matting | 15.4 | 13.4 | 16.9 | 16 |
| Shared Matting | 16.3 | 15.1 | 19 | 14.8 |

(a) SAD

| Comprehensive sampling | 12.7 | 11.9 | 12.4 | 13.8 |
|---|---|---|---|---|
| SVR Matting | 13 | 16.9 | 11.3 | 10.9 |
| Comprehensive Weighted Color and Texture | 13.7 | 13.8 | 14.3 | 13 |
| **LocalSamplingAndKnnClassification** | **14.8** | **16.6** | **12.8** | **14.9** |
| Anonymous SP_Lett_Subm | 15.9 | 16.4 | 12.8 | 18.5 |
| Sparse coded matting | 16.3 | 18.8 | 17.3 | 13 |
| Weighted Color and Texture Matting | 16.4 | 15.3 | 17.4 | 16.5 |

(b) MSE

| Shared Matting | 14.7 | 14.9 | 15.9 | 13.4 |
|---|---|---|---|---|
| Comprehensive Weighted Color and Texture | 16.4 | 16 | 18.6 | 14.5 |
| Anonymous SP_Lett_Subm | 18.6 | 20.8 | 14.1 | 20.9 |
| **LocalSamplingAndKnnClassification** | **19.2** | **20.8** | **18.5** | **18.4** |
| Global Sampling Matting (filter version) | 19.5 | 17.9 | 19.4 | 21.1 |
| LNCLM matting | 20 | 22 | 19.9 | 18 |
| Weighted Color and Texture Matting | 20.5 | 19.6 | 19.8 | 22.1 |

(c) Gradient

| SVR Matting | 14.2 | 16 | 12 | 14.6 |
|---|---|---|---|---|
| Improved color matting | 14.8 | 16.8 | 12.6 | 14.9 |
| anonymous_submission | 15 | 13.6 | 14.8 | 16.6 |
| **LocalSamplingAndKnnClassification** | **16.3** | **19.3** | **15.5** | **14** |
| Local Spline Regression (LSR) | 16.5 | 21.3 | 15.9 | 12.4 |
| Anonymous SP_Lett_Subm | 17.3 | 17.3 | 19.3 | 15.3 |
| Cell-based matting Laplacian | 17.5 | 19.4 | 14.9 | 18.4 |

(d) Connectivity

Table (9) Online rating(2015/12/23)

## 5.4 Limitations

The shortcomings of this method include overdependence on the results of classification, which means that the resulting matting effects will enjoy a great quality when the classification results are accurate, otherwise, the resulting effects of the matting will be negatively affected. If the image's foreground and background overlap, although by constructing different feature spaces can somehow improve the classification accuracy, however, if the image's background and foreground colors overlap and their coordinates are very close, it still will seriously affect the results of the classification, leading to worsening effects of the matting. Images with great propagation do not need extra marks and labels, in which case the CF method has already achieved great and satisfactory effects.

# 6   Conclusion

This paper makes improvements on the limitations of CF matting algorithm when used in complex areas such as hole areas or long and narrow areas, automatically providing certain tags for the unknown areas of the image by local sampling and KNN classifier classification to improve the efficacy of matting in complex areas. When establishing KNN classifier, the corresponding feature spaces are constructed based on the different color components of the image to improve the accuracy of classifier. Based on the sampling performance of the image's unknown areas, adaptively use local sampling and classifier to preliminary process the different unknown areas. Finally when combined with CF calculation, the weight value of constraint item is a changing value based on the previous calculating results, more effective in the preliminary treatment. It shows that on images with good propagation , this method is as effective as the Closed Form method, while on images in complex regions, it can perform even better than Closed Form.

Future research will include: study the feasibility of using other methods for the establishment and use of classifier such as using other distance for the calculation of similarity; establish a more effective image feature space to improve the accuracy of classification; finding a better way to calculate the constraint matrix; using GPU for the parallel acceleration of the algorithm.